%% file: acl_latex.tex
\documentclass[11pt]{article}

\usepackage[final]{acl}

\usepackage{times}
\usepackage{latexsym}

\usepackage[T1]{fontenc}

\usepackage[utf8]{inputenc}

\usepackage{microtype}

\usepackage{inconsolata}

\usepackage{graphicx}

\input{section/package}
\input{section/notation}

\title{Policy and World Modeling Co-Training for Language Agents}

\author{
\textbf{Ning Lu\textsuperscript{1,2,}\thanks{Equal contribution.}},
\textbf{Baijiong Lin\textsuperscript{3,}\hyperlink{Hfootnote.1}{\footnotemark[1]}},
\textbf{Shengcai Liu\textsuperscript{1,}\thanks{Corresponding author: liusc3@sustech.edu.cn.}},
\textbf{Jiahao Wu\textsuperscript{1,4}},
\textbf{Haoze Lv\textsuperscript{1}},
\textbf{Yanbin Wei\textsuperscript{2,5}},
\\
\textbf{Lingting Zhu\textsuperscript{6}},
\textbf{Shengju Qian\textsuperscript{7}},
\textbf{Xin Wang\textsuperscript{7}},
\textbf{Yingcong Chen\textsuperscript{2,3}},
\textbf{Qi Wang\textsuperscript{5}},
\textbf{Ke Tang\textsuperscript{1}}
\\
\parbox{\textwidth}{ \centering \normalsize
  \textsuperscript{1}Guangdong Provincial Key Laboratory of Brain-Inspired Intelligent Computation,\\ Department of Computer Science and Engineering, Southern University of Science and Technology
  \\
  \textsuperscript{2}The Hong Kong University of Science and Technology
  \\
  \textsuperscript{3}The Hong Kong University of Science and Technology (Guangzhou)  \\
  \textsuperscript{4}The Hong Kong Polytechnic University \quad \textsuperscript{5}Southern University of Science and Technology \\
  \textsuperscript{6}{The University of Hong Kong} \quad \textsuperscript{7}LIGHTSPEED
}
}

\begin{document}
\maketitle
\begin{abstract}
Reinforcement learning (RL) improves large language model (LLM) agents by teaching them which actions lead to high rewards, but provides little supervision on what those actions do to the environment.
World modeling (WM) can fill this gap, yet existing approaches often require separate simulators, extra training stages, or additional inference-time computation.
We observe that on-policy RL rollouts already contain the needed signal: each transition pairs an action with its resulting next observation.
Based on this observation, we propose \textbf{\methodname{}}, a \textbf{P}olicy \textbf{a}nd \textbf{W}orld modeling co-training framework that adds auxiliary WM supervision to the same policy during RL, without changing the inference paradigm.
To make auxiliary WM supervision informative and stable, {\methodname{}} introduces three components: action-entropy-based WM data selection, noise-tolerant WM loss, and reward-adaptive loss balancing.
Experiments on three agentic task benchmarks show consistent improvements over strong RL baselines across models and RL algorithms.
These results suggest that standard RL rollouts are a practical source of WM supervision for language-agent training. Our code  is available at \url{https://github.com/ColinLu50/Policy-WorldModel-Agent}.
\end{abstract}

\input{section/introduction}
\input{section/background}

\input{section/method}
\input{section/experiments}

\input{section/related_work}

\input{section/conclusion}

\bibliography{custom,intro}

\appendix

\input{appendix/other_implementation}

\input{appendix/other_main_results}

\input{appendix/other_related_work}

\input{appendix/others}

\end{document}

%% file: section/package.tex
\usepackage[inline, shortlabels]{enumitem}
\usepackage{url}            
\usepackage{booktabs}       
\usepackage{amsfonts}       
\usepackage{nicefrac}       
\usepackage{microtype}      
\usepackage{xcolor}         

\usepackage{bm}
\usepackage{multirow}
\usepackage{pifont}
\usepackage{wrapfig}
\usepackage{colortbl}

\usepackage{amsmath}
\usepackage{amssymb}
\usepackage{mathtools}
\usepackage{amsthm}
\usepackage[most]{tcolorbox}
\usepackage{graphicx}

\usepackage{algorithm}
\usepackage{algpseudocode}
\usepackage{float}

\usepackage{hyperref}
\usepackage[capitalise,noabbrev]{cleveref}

\usepackage{subcaption}
\usepackage{dsfont}
\theoremstyle{plain}

\theoremstyle{definition}

\theoremstyle{remark}

\usepackage{enumitem}

\usepackage{algpseudocode}
\usepackage{float}

\definecolor{midnightgreen}{rgb}{0.0, 0.29, 0.33}
\definecolor{deepgreen}{HTML}{055c29}
\definecolor{deeppurple}{HTML}{7030a0}
\definecolor{deepblue}{HTML}{171d91}
\definecolor{brown}{HTML}{843c0c}
\definecolor{shadered}{HTML}{ffe5e5}
\definecolor{shadegreen}{HTML}{e5f7ed}
\definecolor{msftBlack}{RGB}{0,0,0}
\definecolor{lightred}{RGB}{255,163,163}
\definecolor{deepred}{RGB}{153,0,0}

\definecolor{barblue}{RGB}{90,120,180}
\definecolor{barorange}{RGB}{225,124,5}

\newcommand{\methodname}{PaW}

\definecolor{softblue}{RGB}{30, 90, 160}
\hypersetup{
  colorlinks=true,
  linkcolor=deepred,
  citecolor=softblue,
  urlcolor=magenta
}

\usepackage{xspace}

%% file: section/notation.tex
\newcommand{\vo}{\bm{o}}
\newcommand{\va}{\bm{a}}
\newcommand{\vh}{\bm{h}}

\newcommand{\vtau}{\bm{\tau}}
\newcommand{\vtheta}{\bm{\theta}}
\newcommand{\vphi}{\bm{\phi}}

\newcommand{\vpitheta}{\bm{\pi_\theta}}
\newcommand{\vpiphi}{\bm{\pi_\phi}}

\newcommand{\evaltok}[1]{%
  \begingroup
  \setlength{\fboxsep}{1pt}%
  \colorbox{gray!25}{\textbf{\texttt{#1}}}%
  \endgroup
}

%% file: section/introduction.tex
\section{Introduction}
\label{sec:intro}

Reinforcement learning (RL) has become a dominant paradigm for improving large language model (LLM) agent performance ~\cite{deepseekai2026deepseekv4,glm5team2026glm5vibecodingagentic}.
However, standard RL optimizes actions for reward maximization without learning their consequences, leaving agents brittle to invalid operations, irreversible state changes, and delayed failures in long-horizon tasks~\cite{hao2023rap,liu2026imaginethenplan}.
World modeling (WM) addresses this gap by predicting the next observation from the interaction history and the chosen action, encouraging the agent to internalize environment dynamics rather than memorize which actions get rewards~\cite{zhang2025earlyexperience}.

\begin{figure}[t]
  \includegraphics[width=\columnwidth]{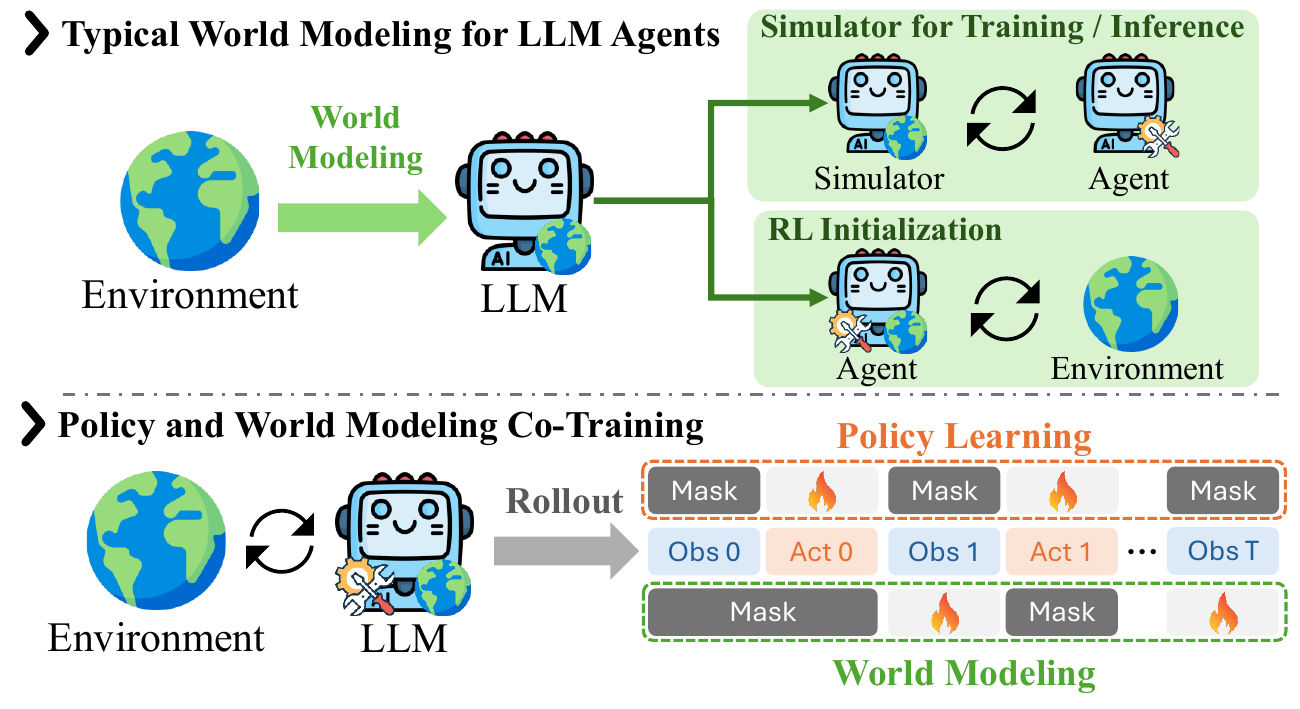}
  \vspace{-0.3in}
  \caption{Comparison of world modeling paradigms for LLM agents. While prior methods rely on separate simulators, additional training, or inference-time planning, our \methodname{} jointly optimizes policy learning and world modeling within the same model.}
  \label{fig:wm_methods}
\end{figure}

Existing WM methods for LLM agents typically introduce this ability outside the standard RL training, as shown in \cref{fig:wm_methods}.
One line of work trains a world model simulator, either as a separate model or within the policy model itself, to generate imagined trajectories for RL training or to scale inference-time planning~\cite{gu2025webdreamer,fang2025webevolver,xiao2026webworld,li2026wordworldlargelanguage,liu2026imaginethenplan}.
Another line of work first instills WM ability into the model and then fine-tunes it with RL training~\cite{zhang2025earlyexperience,yu2026rwml}.
In both cases, WM learning incurs extra cost: a separate model, an additional training stage, or inference-time computation.
This raises a question: \emph{can world-modeling ability be learned jointly with policy improvement within the same RL training process?}

\begin{figure*}[t]
\centering
\captionsetup[subfigure]{labelformat=parens,labelsep=none}

\begin{subfigure}[t]{0.245\textwidth}
\centering
\includegraphics[width=\linewidth]{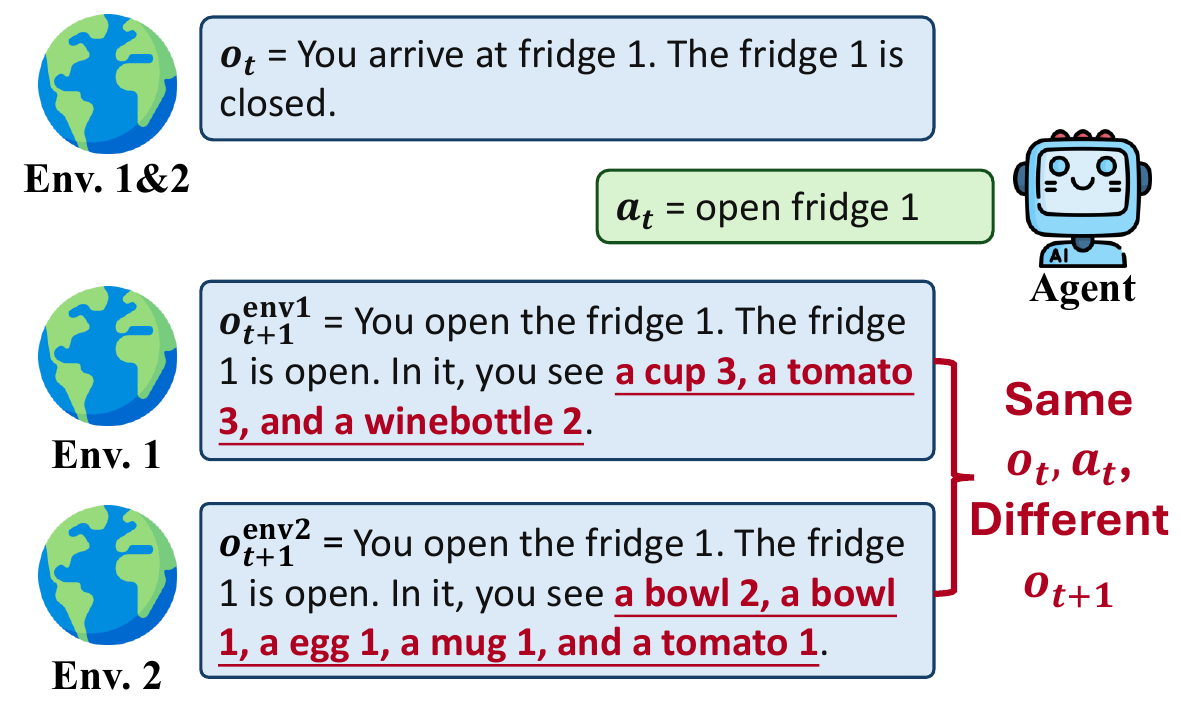}
\caption{}
\label{fig:cmae-noise-alfworld}
\end{subfigure}
\hfill
\begin{subfigure}[t]{0.245\textwidth}
\centering
\includegraphics[width=\linewidth]{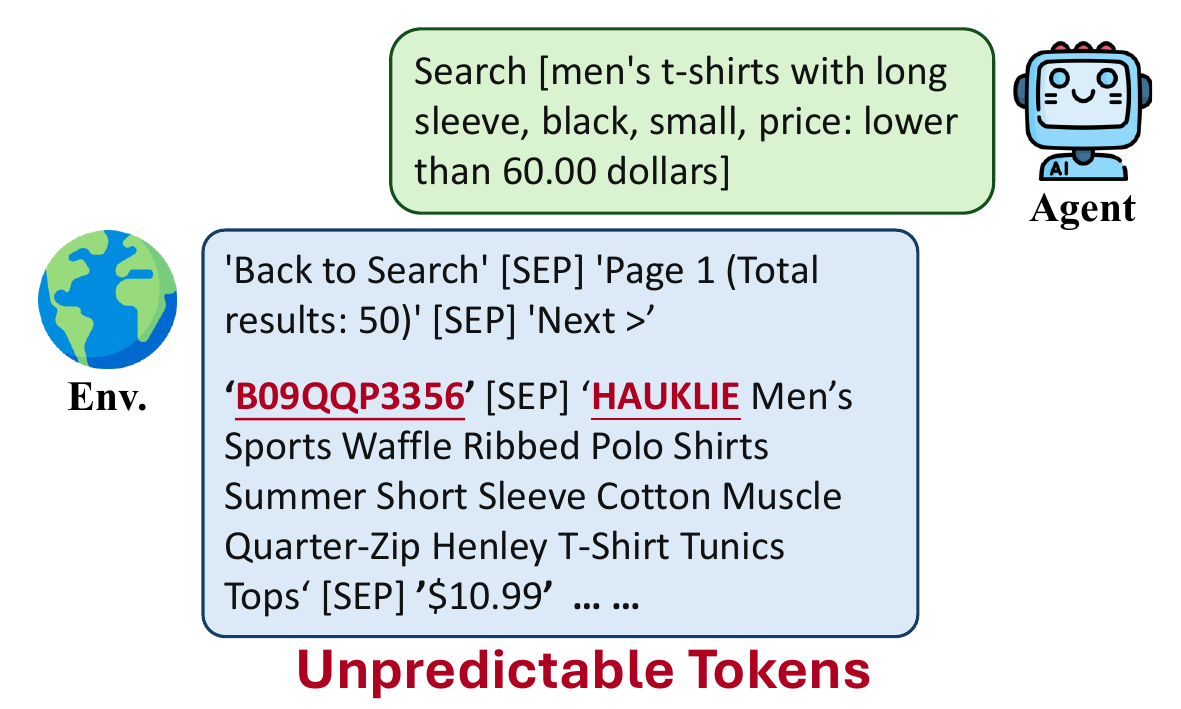}
\caption{}
\label{fig:cmae-noise-webshop}
\end{subfigure}
\hfill
\begin{subfigure}[t]{0.245\textwidth}
\centering
\includegraphics[width=\linewidth]{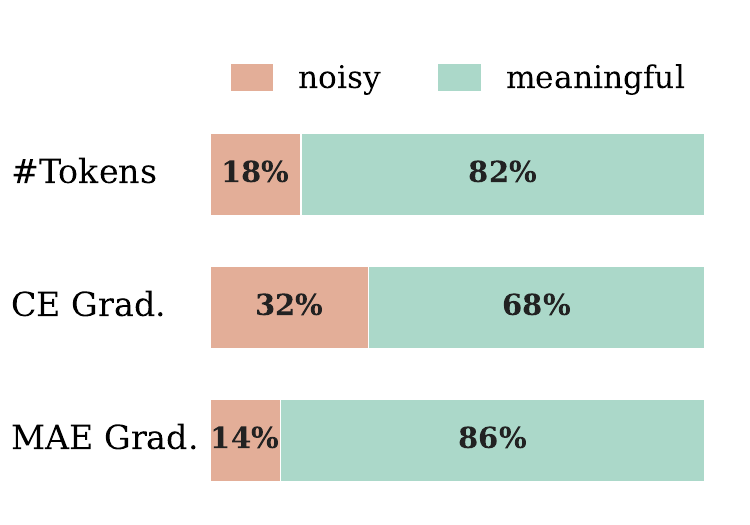}
\caption{}
\label{fig:cmae-grad-share}
\end{subfigure}
\hfill
\begin{subfigure}[t]{0.245\textwidth}
\centering
\includegraphics[width=\linewidth]{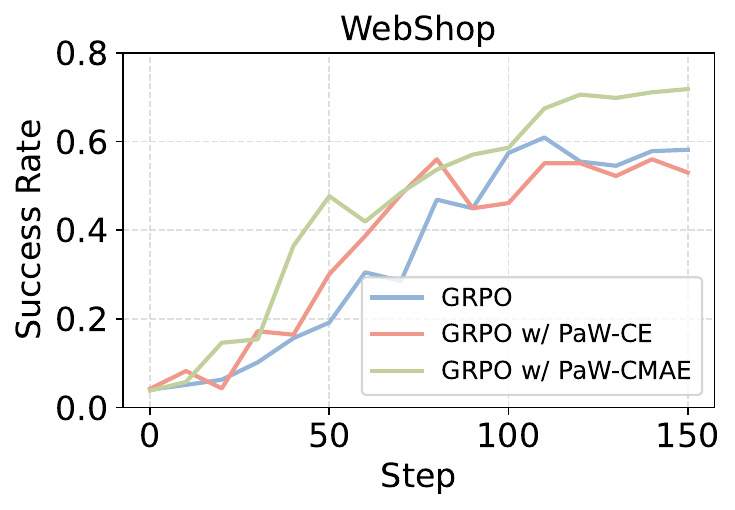}
\caption{}
\label{fig:cmae-webshop-curve}
\end{subfigure}
\vspace{-0.1in}
\caption{
\textbf{Illustration of noisy observation tokens and the effect of clipped MAE loss.}
(a) and (b) show two noisy WM training examples from ALFWorld and WebShop, where the same $(\vo_t, \va_t)$ pair can lead to different next observation $\vo_{t+1}$ in (a) and observations may contain random surface noise in (b).
(c) shows that CE WM loss (\cref{eq:wm-ce}) assigns a disproportionately large gradient share to noisy tokens. See \cref{apx:noise_gradient_analysis} for details.
(d) shows that replacing CE (\cref{eq:wm-ce}) with clipped MAE (\cref{eq:cmae-loss}) during training improves RL performance.
}
\label{fig:cmae-visualization}
\end{figure*}

Our key observation is that on-policy RL rollouts already provide world-modeling supervision. Each interaction step yields both policy supervision from the action and its advantage, and dynamics supervision from the resulting next observation, which reveals what the action causes. While standard RL uses only the former, we exploits the latter as dense action-conditioned next-observation supervision, without requiring additional rollouts.

Motivated by this observation, we propose \textbf{\methodname{}}, a framework for \textbf{P}olicy \textbf{a}nd \textbf{W}orld modeling co-training during on-policy RL post-training.
As shown in \cref{fig:wm_methods,fig:overview}, {\methodname{}} reuses RL rollouts by appending next-observation tokens and applying an auxiliary next-token-prediction loss to train the same model.
Policy learning is unchanged because causal attention prevents later next-observation tokens from affecting the action logits.
In inference, the agent behaves like a standard policy model, with no additional simulation steps.

However, as illustrated in \cref{fig:cmae-noise-alfworld,fig:cmae-noise-webshop}, rollout observations provide noisy supervision: some transitions are uninformative, some target tokens are unpredictable, and the auxiliary WM loss must balance with the RL loss. 
Thus, {\methodname{}} combines three key designs: action-entropy-based WM data selection, clipped mean absolute error (MAE;\citet{GhoshKS2017mae}) loss for noisy observations, and reward-adaptive loss balancing. Together, they make auxiliary WM supervision both informative and stable.
We evaluate {\methodname{}} on two types of agentic tasks: interactive decision-making (ALFWorld and WebShop) and search-augmented QA.
Across models and RL algorithms, {\methodname{}} consistently improves strong RL baselines, with negligible training overhead.
These results suggest that standard RL rollouts already provide a practical source of world-model supervision for language-agent training.

Our contributions are summarized as follows:
 \begin{enumerate*}[(i), series = tobecont, itemjoin = \quad]
    \item We identify next observations in on-policy rollouts as an overlooked source of action-conditioned WM supervision for language-agent RL;
    \item We propose \textbf{\methodname{}}, the first policy and world-modeling co-training method for RL. It reuses on-policy rollouts for joint policy optimization and world-modeling supervision, with high-action-entropy transition selection, clipped MAE loss, and adaptive WM loss balancing;
    \item We show consistent improvements over strong RL baselines on three agentic tasks across models and RL algorithms.
\end{enumerate*}

%% file: section/background.tex
\section{Preliminaries}
\label{sec:prelim}

\paragraph{Problem setup.}
We consider language-agent tasks where a policy $\vpitheta$ solves a user-specified goal through multi-turn interaction with an environment.
At turn $t$, the agent observes $\vo_t$, forms a decision context $\vh_t$ from the instruction and interaction history, and samples a textual or serialized action $\va_t\sim\vpitheta(\cdot\mid\vh_t)$.
The environment executes $\va_t$ and returns reward $r_t$ and next observation $\vo_{t+1}$, yielding a trajectory $\vtau=\{\vo_0,\va_0,r_0,\vo_1,\ldots,\vo_{T-1},\va_{T-1},r_{T-1},\vo_T\}$ with return $R(\vtau)=\sum_{t=0}^{T-1}r_t$.

\begin{figure*}[t]
  \includegraphics[width=\textwidth]{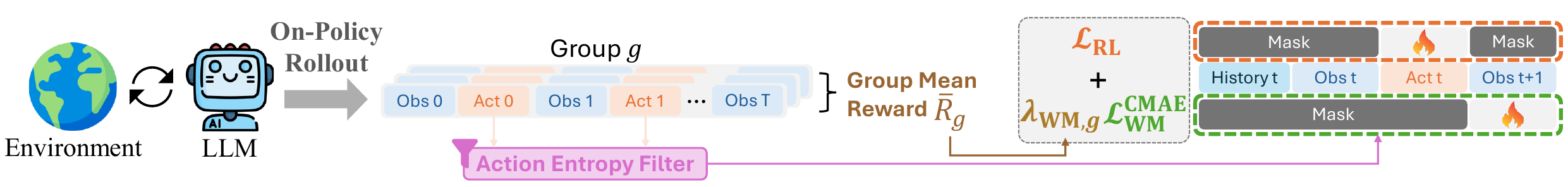}
  \vspace{-0.3in}
  \caption{\textbf{Overview of \methodname{}.} {\methodname{}} introduces auxiliary world modeling to agentic RL via action-entropy WM data selection, clipped MAE, and reward-adaptive loss balancing.}
  \label{fig:overview}
\end{figure*}

\paragraph{On-policy Agentic RL.}
Agentic RL fine-tunes $\vpitheta$ to maximize $J(\vtheta)=\mathbb{E}_{\vtau\sim\vpitheta}[R(\vtau)]$ using sampled rollouts.
Most on-policy algorithms can be abstracted as minimizing an advantage-weighted action loss:
\begin{equation}
\mathcal{L}_{\mathrm{RL}}(\vtheta)
=
-\mathbb{E}_{\vtau}
\left[
\sum_{t=0}^{T-1}
A_t \log \vpitheta(\va_t \mid \vh_t)
\right],
\label{eq:policy-loss}
\end{equation}
where $A_t$ is a reward-derived advantage and $\log \vpitheta(\va_t\mid\vh_t)$ sums token-level log-likelihoods of the action sequence.
Different algorithms, such as GRPO~\citep{shao2024deepseekmath} and GIGPO~\citep{feng2025group}, mainly differ in how they estimate $A_t$ and instantiate the surrogate objective.
Our method is orthogonal to these choices and augments this base RL loss with world modeling supervision.

\paragraph{World-modeling for language agents.}
World modeling aims to capture action-conditioned dynamics by predicting the environment response after an action~\cite{zhang2025earlyexperience}.
For language agents, this becomes next-observation prediction: given $(\vh_t,\va_t)$, an autoregressive model predicts the textual observation $\vo_{t+1}$ with objective:
\begin{equation}
\mathcal{L}_{\mathrm{WM}}(\vphi)
=
-\mathbb{E}
\left[
\log \vpiphi(\vo_{t+1} \mid \vh_t,\va_t)
\right],
\label{eq:prelim-wm}
\end{equation}
where the likelihood is computed over observation tokens.
Learning world modeling enables agents to better understand action outcomes and make better decisions in long-horizon  tasks~\cite{zhang2025earlyexperience,liu2026imaginethenplan}.

%% file: section/method.tex
\section{Methodology}
\label{sec:method}

In this section, we present \textbf{\methodname{}}, a framework for co-training policy learning with world modeling within a single policy model during on-policy RL by augmenting the base RL objective with an auxiliary WM objective:
\begin{equation}
\mathcal{L}_{\mathrm{\methodname{}}}(\vtheta)
=
\mathcal{L}_{\mathrm{RL}}(\vtheta)
+
\lambda_{\mathrm{WM}}\mathcal{L}_{\mathrm{WM}}(\vtheta).
\label{eq:schematic-cotraining}
\end{equation}
Here, both terms update the same parameters $\vtheta$: the RL loss improves action selection, while the WM loss trains next-observation prediction from rollout transitions. 
See \cref{fig:overview} for the overview.
The rest of this section instantiates this objective by describing how we construct WM supervision with action-entropy-based data selection in \cref{sec:method:overview}, make observation prediction robust with clipped MAE in \cref{sec:method:cmae}, set $\lambda_{\mathrm{WM}}$ as a reward-adaptive coefficient in \cref{sec:method:coef}, and summarize training and inference in \cref{sec:method:training}.

\subsection{Constructing World Modeling Supervision from RL Rollouts}
\label{sec:method:overview}

At each RL update, we sample task instances and collect rollout groups $\mathcal{G}$ with the current policy, where each group $g\in\mathcal{G}$ contains trajectories from the same task.
Each group forms a transition pool $\mathcal{P}_g=\{(\vh_t,\va_t,r_t,\vo_{t+1}) \mid \vtau\in g,\; 0\leq t<T\}$, and the update-level pool is $\mathcal{P}=\bigcup_{g\in\mathcal{G}}\mathcal{P}_g$.

Each transition in this pool serves two roles: $(\vh_t,\va_t,A_t)$, with $A_t$ computed from rewards, provides the action-level signal for policy learning, while $(\vh_t,\va_t,\vo_{t+1})$ provides the token-level target for next-observation prediction.
However, using all transitions for WM learning can overweight redundant or low-informative action consequences.
Thus, we select a subset of transitions for the auxiliary next-observation loss.

\paragraph{WM data selection.}
For each transition, let $q_{t,i}=\vpitheta\left(\va_t^{(i)}\mid\vh_t,\va_t^{(<i)}\right)$ denote the next-token probability at the $i$-th action position.
We compute a sample-based estimate of the action-token entropy using the realized action tokens:
$
H(\va_t\mid\vh_t)
=
-\frac{1}{|\va_t|}
\sum_{i=1}^{|\va_t|}
\log q_{t,i},
$
a Monte Carlo estimate over the sampled action sequence.
Intuitively, high-entropy actions correspond to decision points where the policy assigns probability mass to more diverse action alternatives.
Their resulting observations are therefore more informative for learning action-conditioned environment transitions than highly deterministic, repetitive actions.
We apply the selector to all candidate WM transitions in the current RL update.
Given a retained fraction $\alpha\in(0,1]$, we keep the top-$\alpha$ fraction by action entropy:
\begin{equation}
\mathcal{S}_{\alpha}
=
\operatorname{Top}_{\alpha}\left(\mathcal{P};\,H(\va_t\mid\vh_t)\right),
\label{eq:entropy-selection}
\end{equation}
where $\operatorname{Top}_{\alpha}$ returns the highest-entropy $\alpha$ fraction of $\mathcal{P}$.
The RL loss is computed on all generated actions in the rollout groups, while the world-model loss is computed only on selected transitions in $\mathcal{S}_{\alpha}$.

\paragraph{Co-training in one forward pass.}
Operationally, each transition is serialized as $(\vh_t,\va_t,\vo_{t+1})$, with rewards used only for advantage computation.
We apply an action-token mask for the base RL loss and an observation-token mask for the auxiliary loss on appended next-observation tokens in $\mathcal{S}_{\alpha}$.
Causal attention prevents action tokens from attending to appended observations, so $\vo_{t+1}$ does not affect action logits; meanwhile, observation logits are conditioned on $(\vh_t,\va_t)$ for next-observation prediction.
The entropy uses action-token distributions already available in the rollout or training pass, introducing no additional model forward.

\subsection{Clipped MAE Loss for Noisy Observation Prediction}
\label{sec:method:cmae}

After action-entropy-based selection, the selected transitions provide action-conditioned targets for world modeling.
A direct instantiation is to apply the cross-entropy objective from \cref{eq:prelim-wm} to the next-observation tokens.
For a selected transition $(\vh_t,\va_t,r_t,\vo_{t+1})\in\mathcal{S}_{\alpha}$, let $p_{t,i}=\vpitheta\!\left(\vo_{t+1}^{(i)}\mid\vh_t,\va_t,\vo_{t+1}^{(<i)}\right)$ be the probability assigned to the $i$-th target token in $\vo_{t+1}$.
The standard CE loss on the selected set can be written as:
\begin{equation}
\mathcal{L}_{\mathrm{WM}}^{\mathrm{CE}}(\vtheta;\mathcal{S}_{\alpha})
=
\mathbb{E}_{\mathcal{S}_{\alpha}}
\left[-\frac{1}{|\vo_{t+1}|}\sum_{i=1}^{|\vo_{t+1}|}\log p_{t,i}\right].
\label{eq:wm-ce}
\end{equation}
Although CE is standard for language modeling, it is poorly matched to observation prediction in agentic environments.
As illustrated in \cref{fig:cmae-noise-alfworld,fig:cmae-noise-webshop}, next observations may be stochastic, non-unique, or contaminated by nuisance tokens such as IDs, product names, and random strings.
Overfitting these tokens can consume optimization capacity without improving action-relevant dynamics.

\paragraph{MAE-style token loss.}
To reduce the influence of hard and low-probability observation tokens, we replace the per-token CE loss $\ell_{\mathrm{CE}}^{(i)}=-\log p_{t,i}$ with a mean MAE loss $\ell_{\mathrm{MAE}}^{(i)}=1-p_{t,i}$~\cite{GhoshKS2017mae}.
Let $z_{t,i}$ denote the logit assigned to the target token.
The magnitudes of the corresponding target-logit gradients are:
\begin{equation}
\left|
\frac{\partial \ell_{\mathrm{CE}}^{(i)}}{\partial z_{t,i}}
\right|
=
1-p_{t,i},
\quad
\left|
\frac{\partial \ell_{\mathrm{MAE}}^{(i)}}{\partial z_{t,i}}
\right|
=
p_{t,i}(1-p_{t,i}).
\label{eq:ce-mae-gradient}
\end{equation}
For a low-probability target token, the CE magnitude approaches one, whereas the MAE magnitude decreases proportionally with $p_{t,i}$.
This makes MAE less sensitive to unpredictable observation fragments, as also reflected by the gradient-share analysis in \cref{fig:cmae-grad-share}.
This loss enables WM learning, as shown in \cref{apx:wm_learn}.

\paragraph{Confidence clipping.}
MAE alone still keeps optimizing tokens after the model already predicts them with high confidence.
However, in stochastic textual environments, forcing the model to further fit one observed realization may encourage memorization of arbitrary surface details.
We therefore introduce a token-level confidence mask:
\begin{equation}
m_{t,i}
=
\mathds{1}\!\left[p_{t,i}\leq\rho\right],
\label{eq:cmae-mask}
\end{equation}
where $\rho$ is a confidence threshold.
Tokens with $p_{t,i}>\rho$ are treated as sufficiently learned and are removed from the auxiliary target.
For one selected transition, the clipped MAE loss is:
\begin{equation}
\ell_{\mathrm{CMAE}}(\vh_t,\va_t,\vo_{t+1})
=
\frac{
\sum_{i=1}^{|\vo_{t+1}|}m_{t,i}(1-p_{t,i})
}{
|\vo_{t+1}|
}.
\label{eq:cmae-transition-loss}
\end{equation}
The WM loss on the selected transitions is then:
\begin{equation}
\mathcal{L}_{\mathrm{WM}}^{\mathrm{CMAE}}(\vtheta;\mathcal{S}_{\alpha})
=
\mathbb{E}_{\mathcal{S}_{\alpha}}
\left[\ell_{\mathrm{CMAE}}(\vh_t,\va_t,\vo_{t+1})\right].
\label{eq:cmae-loss}
\end{equation}
This loss focuses learning on insufficiently predicted observation tokens, while avoiding the excessive CE pressure on noisy or non-action-relevant tokens.
Examples are shown in \cref{apx:cmae_token_analysis}.

\subsection{Reward-Adaptive Loss Balancing}
\label{sec:method:coef}

Even after entropy filtering and token-level clipping, WM supervision remains dense because every selected observation can contribute many token-level gradients.
If applied with a fixed large weight, this auxiliary objective may dominate the sparse reward-driven policy update.
Moreover, the need for auxiliary dynamics learning is task-dependent: low-performing rollout groups can benefit more from additional next-observation supervision, while high-performing groups should focus more on refining the policy objective.

We therefore instantiate the schematic coefficient $\lambda_{\mathrm{WM}}$ in \cref{eq:schematic-cotraining} as a reward-adaptive coefficient for each rollout group $g\in\mathcal{G}$.
We set:
\begin{equation}
\lambda_{\mathrm{WM},g}
=
\left[
1-\frac{\bar{R}_g}{R_{\mathrm{max}}}
\right]_+,
\label{eq:group-coef}
\end{equation}
where where $[\cdot]_+$ denotes the positive-part operator, $\bar{R}_g=|g|^{-1}\sum_{\vtau\in g}R(\vtau)$ denotes its mean episode return and $R_{\mathrm{max}}$ denotes the maximum attainable episode return in the environment.
Low-return groups receive stronger WM supervision, whereas as $\bar{R}_g$ approaches $R_{\mathrm{max}}$, $\lambda_{\mathrm{WM},g}$ decreases, allowing the base RL objective to dominate.

\begin{algorithm}[t]
\small
\caption{On-policy RL training with \methodname{}.}
\label{alg:wm-step}
\begin{algorithmic}[1]
\For{each RL update}
    \State Sample tasks and collect rollout groups $\mathcal{G}$ using $\vpitheta$;
    \State Form transition pools $\{\mathcal{P}_g\}_{g\in\mathcal{G}}$ and $\mathcal{P}=\bigcup_g\mathcal{P}_g$;
    \State Compute selected transitions $\mathcal{S}_{\alpha}$ using \cref{eq:entropy-selection};
    \For{each rollout group $g\in\mathcal{G}$}
        \State Set $\mathcal{S}_{\alpha,g}:=\mathcal{S}_{\alpha}\cap\mathcal{P}_g$;
        \State Compute $\mathcal{L}_{\mathrm{RL}}(\vtheta;g)$ using \cref{eq:policy-loss};
        \State Compute $\mathcal{L}_{\mathrm{WM}}^{\mathrm{CMAE}}(\vtheta;\mathcal{S}_{\alpha,g})$ using \cref{eq:cmae-loss};
        \State Compute $\lambda_{\mathrm{WM},g}$ using \cref{eq:group-coef};
    \EndFor
    \State Optimize \cref{eq:paw-objective} to update $\vtheta$;
\EndFor
\end{algorithmic}
\end{algorithm}

\begin{table*}[t]
\centering
\resizebox{0.9\textwidth}{!}{
\begin{tabular}{llccccccc|cc}
\toprule
\multirow{2}{*}{Type} & \multirow{2}{*}{Method} & \multicolumn{7}{c|}{\textbf{ALFWorld}} & \multicolumn{2}{c}{\textbf{WebShop}} \\
 & & Pick & Look & Clean & Heat & Cool & Pick2 & Avg. & Score & Succ.\\
\midrule
\multicolumn{11}{l}{\textit{Closed-Source Model}} \\
Prompting& GPT-4o & 75.3 & 60.8 & 31.2 & 56.7 & 21.6 & 49.8 & 48.0 & 31.8 & 23.7 \\
Prompting& Gemini-2.5-Pro & 92.8 & 63.3 & 62.1 & 69.0 & 26.6 & 58.7 & 60.3 & 42.5 & 35.9 \\
\midrule
\multicolumn{11}{l}{\textit{Qwen2.5-1.5B-Instruct}} \\
Prompting& Qwen2.5 & 5.9 & 5.5 & 3.3 & 9.7 & 4.2 & 0.0 & 4.1 & 23.1 & 5.2 \\
Prompting& ReAct & 17.4 & 20.5 & 15.7 & 6.2 & 7.7 & 2.0 & 12.8 & 40.1 & 11.3 \\
Prompting& Reflexion & 35.3 & 22.2 & 21.7 & 13.6 & 19.4 & 3.7 & 21.8 & 55.8 & 21.9 \\
RL Training& GRPO & 86.5 & 46.3 & 79.0 & 70.2 & 69.1 & 47.8 & 70.0 & 75.6 & 60.6 \\
\rowcolor{gray!25}RL Training& \textbf{GRPO w/ \methodname{}} & \textbf{87.8} & \textbf{59.3} & \textbf{84.5} & \textbf{73.7} & \textbf{75.4} & \textbf{69.6} & \textbf{77.9} & \textbf{83.8} & \textbf{68.6} \\
RL Training& GIGPO & 95.3 & \textbf{84.3} & 87.7 & \textbf{92.6} & 79.8 & 82.3 & 87.6 & 83.2 & 66.2 \\
\rowcolor{gray!25}RL Training& \textbf{GIGPO w/ \methodname{}} & \textbf{95.3} & 83.3 & \textbf{91.8} & 89.5 & \textbf{89.1} & \textbf{84.5} & \textbf{90.4} & \textbf{87.7} & \textbf{75.3} \\
\midrule
\multicolumn{11}{l}{\textit{Qwen2.5-7B-Instruct}} \\
Prompting& Qwen2.5 & 33.4 & 21.6 & 19.3 & 6.9 & 2.8 & 3.2 & 14.8 & 26.4 & 7.8 \\
Prompting& ReAct & 48.5 & 35.4 & 34.3 & 13.2 & 18.2 & 17.6 & 31.2 & 46.2 & 19.5 \\
Prompting& Reflexion & 62.0 & 41.6 & 44.9 & 30.9 & 36.3 & 23.8 & 42.7 & 58.1 & 28.8 \\
RL Training& GRPO & 90.8 & 66.1 & \textbf{89.3} & 74.7 & 72.5 & 64.7 & 77.6 & 75.4 & 66.5 \\
\rowcolor{gray!25}RL Training& \textbf{GRPO w/ \methodname{}} & 90.4 & \textbf{80.7} & 86.8 & \textbf{82.9} & \textbf{76.5} & \textbf{67.3} & \textbf{80.6} & \textbf{84.5} & \textbf{70.5} \\
RL Training& GIGPO & 97.7 & 82.7 & \textbf{98.8} & 83.7 & 89.3 & 79.2 & 90.8 & 85.0 & 73.8 \\
\rowcolor{gray!25}RL Training& \textbf{GIGPO w/ \methodname{}} & \textbf{98.2} & \textbf{85.6} & 98.6 & \textbf{84.5} & \textbf{91.5} & \textbf{84.3} & \textbf{91.8} & \textbf{87.6} & \textbf{76.7} \\
\bottomrule
\end{tabular}
}
\caption{Performance on ALFWorld and WebShop. For ALFWorld, we report the success rate (\%) for each subtask and the overall average. For WebShop, we report the average score and success rate (\%). Bold numbers indicate the better result between each vanilla RL baseline and its \methodname{}-augmented variant. Full results with standard variance can be found in \cref{apx:full_results_alfworld_webshop}.}
\label{tab:main}
\end{table*}

\subsection{Training and Inference}
\label{sec:method:training}

Combining the three designs above, the co-training objective of \methodname{} in \cref{eq:schematic-cotraining} becomes the following final objective, which preserves the base on-policy RL loss over each rollout group while adding a reward-adaptively weighted CMAE world-model loss on the entropy-selected transitions:
\begin{equation}
\begin{aligned}
\mathcal{L}_{\mathrm{\methodname{}}}(\vtheta)
=
&\mathbb{E}_{g\in\mathcal{G}}
\Big[
\mathcal{L}_{\mathrm{RL}}(\vtheta;g) \\
&+\lambda_{\mathrm{WM},g}\mathcal{L}_{\mathrm{WM}}^{\mathrm{CMAE}}
(\vtheta;\mathcal{S}_{\alpha,g})
\Big].
\end{aligned}
\label{eq:paw-objective}
\end{equation}
where $\mathcal{S}_{\alpha,g}=\mathcal{S}_{\alpha}\cap\mathcal{P}_g$, $\mathcal{L}_{\mathrm{RL}}(\vtheta;g)$ is the base on-policy RL loss from \cref{eq:policy-loss}, and the auxiliary loss is computed only on $\mathcal{S}_{\alpha,g}$.
If a group contains no selected transition, its auxiliary term is omitted.

\cref{alg:wm-step} summarizes one on-policy RL update with \methodname{}.
The procedure follows the base RL algorithm for rollout collection and advantage computation, then co-trains the same model with globally selected observation supervision and group-specific auxiliary weighting. Therefore, \methodname{} requires no additional environment interaction, no separate model, and no separate training stage.
It also introduces no additional model forward: the action entropy, action-token loss, and observation-token loss are computed from distributions already available during rollout generation or the masked training pass.

After training, \methodname{} introduces no inference-time change.
The model receives the decision context $\vh_t$ and generates the next action $\va_t$ exactly as a standard policy model.
It does not rollout imagined observations, perform planning with a simulator, or call an additional model.
Thus, all benefits of WM co-training are obtained during training, while deployment keeps the same interface and cost as the underlying RL-trained agent.

%% file: section/experiments.tex
\begin{table*}[t]
\centering
\resizebox{0.9\textwidth}{!}{
\begin{tabular}{llcc@{\,\,\,}c|ccc@{\,\,\,}c|c}
\toprule
\multirow{2}{*}{Type} & \multirow{2}{*}{Method} & \multicolumn{3}{c|}{\textbf{Single-Hop QA}} & \multicolumn{4}{c|}{\textbf{Multi-Hop QA}} & \multirow{2}{*}{Avg.}\\
& & NQ$^{\dagger}$ & TriviaQA$^{\star}$ & PopQA$^{\star}$ & HotpotQA$^{\dagger}$ & 2Wiki$^{\star}$ & MuSiQue$^{\star}$ & Bamboogle$^{\star}$ \\
\midrule
\multicolumn{10}{l}{\textit{Qwen2.5-3B-Instruct}} \\
RL Training & Search-R1  & 34.1 & 54.5 & 37.8 & 32.4 & 31.9 & 10.3 & 26.4 & 32.5 \\
RL Training & ZeroSearch  & 41.4 & 57.4 & 44.8 & 27.4 & 30.0 & 9.8 & 11.1 & 31.7 \\
RL Training & GRPO  & 44.9 & 60.9 & 46.1 & 37.9 & 39.5 & 13.7 & 64.1 & 43.9 \\
\rowcolor{gray!25}RL Training & \textbf{GRPO w/ \methodname{}}  & \textbf{45.8} & \textbf{61.2} & \textbf{47.5} & \textbf{39.4} & \textbf{40.1} & \textbf{14.4} & \textbf{65.2} & \textbf{44.8} \\
RL Training & GIGPO  & 42.5 & 58.5 & 46.3 & 35.2 & 34.4 & 11.8 & 60.0 & 41.2 \\
\rowcolor{gray!25}RL Training & \textbf{GIGPO w/ \methodname{}}  & \textbf{46.2} & \textbf{61.8} & \textbf{46.7} & \textbf{37.6} & \textbf{38.0} & \textbf{13.9} & \textbf{64.9} & \textbf{44.2} \\
\midrule
\multicolumn{10}{l}{\textit{Qwen2.5-7B-Instruct}} \\
RL Training & Search-R1  & 39.3 & 61.0 & 39.7 & 37.0 & 40.1 & 14.6 & 36.8 & 38.4 \\
RL Training & ZeroSearch  & 43.6 & 61.8 & 51.5 & 34.6 & 35.2 & 18.4 & 27.8 & 39.0 \\
RL Training & GRPO  & 47.9 & 63.9 & 47.8 & 43.9 & 43.6 & 18.3 & 69.6 & 47.9 \\
\rowcolor{gray!25}RL Training & \textbf{GRPO w/ \methodname{}}  & \textbf{48.9} & \textbf{64.9} & \textbf{48.5} & \textbf{44.9} & \textbf{45.1} & \textbf{18.9} & \textbf{70.1} & \textbf{48.8} \\
RL Training & GIGPO  & 46.1 & 64.4 & 46.0 & 40.2 & 41.2 & 16.4 & 68.9 & 45.8 \\
\rowcolor{gray!25}RL Training & \textbf{GIGPO w/ \methodname{}}  & \textbf{46.5} & \textbf{66.0} & \textbf{47.2} & \textbf{42.2} & \textbf{42.8} & \textbf{18.6} & \textbf{69.5} & \textbf{47.5} \\
\bottomrule
\end{tabular}
}
\vspace{-0.05in}
\caption{Performance on search-augmented QA tasks. Agents are trained on NQ and HotpotQA. $\dagger$ and $\star$ indicate in-domain and out-of-domain evaluation datasets, respectively. Avg. denotes the average score across all seven benchmarks. Bold indicates the better result between each vanilla RL baseline and its \methodname{}-augmented variant.}
\label{tab:main_qa}
\end{table*}

\section{Experiments}
In this section, we empirically evaluate \methodname{} on two types of agentic tasks, including interactive decision-making (i.e., ALFWorld and WebShop) and search-augmented QA. 

\subsection{Experiment Setup}
\paragraph{Benchmarks.}
We evaluate \methodname{} on interactive decision-making and search-augmented QA tasks.
For interactive decision-making, we use ALFWorld~\cite{shridhar2020alfworld}, an embodied text-based environment with 3,827 household tasks instances across six household task categories: Pick \& Place (Pick), Examine in Light (Look), Clean \& Place (Clean), Heat \& Place (Heat), Cool \& Place (Cool), and Pick Two \& Place (Pick2), and WebShop~\cite{yao2022webshop}, a web shopping environment with over 110K products and 12K user instructions.
For search-augmented QA, we evaluate multi-turn tool use on single-hop benchmarks including NQ~\cite{kwiatkowski2019natural}, TriviaQA~\cite{joshi2017triviaqa}, and PopQA~\cite{mallen2022not}, as well as multi-hop benchmarks including HotpotQA~\cite{yang2018hotpotqa}, 2Wiki~\cite{ho2020constructing}, MuSiQue~\cite{trivedi2022musique}, and Bamboogle~\cite{press2022measuring}.

\paragraph{Baselines.}
We use GRPO~\cite{shao2024deepseekmath} and GIGPO~\cite{feng2025group} as the main RL baselines, and compare each vanilla algorithm with its \methodname{}-augmented variant.
For ALFWorld and WebShop, we also compare against closed-source LLM agents, including GPT-4o~\cite{achiam2023gpt} and Gemini-2.5-Pro~\cite{team2023gemini}, and prompting-based agents, including ReAct~\cite{yao2023react} and Reflexion~\cite{shinn2024reflexion}.
For search-augmented QA, we additionally include Search-R1~\cite{jin2025search} and ZeroSearch~\cite{sun2025zerosearch} as representative RL-based search-agent baselines.

\paragraph{Implementation details.}
Following \cite{feng2025group}, we use Qwen2.5-1.5B/7B-Instruct~\cite{yang2024qwen2} for ALFWorld and WebShop, and Qwen2.5-3B/7B-Instruct for search-augmented QA.
We set the rollout group size to $8$ for ALFWorld and WebShop and $5$ for search-augmented QA, where E5~\cite{wang2022text} is used as the retriever with at most $4$ interaction turns.
For \methodname{}, we set entropy selection ratio to $\alpha=0.75$ and the clipping threshold to $\rho=0.2$, while keeping all RL hyperparameters identical to the corresponding vanilla RL baseline.
All results are averaged over three random seeds. More details are shown in \cref{apx:implementation}.

\begin{table}[t]
\centering
\resizebox{0.9\columnwidth}{!}{
\begin{tabular}{l l cc}
\toprule
RL & Model & Vanilla & w/ \methodname{} \\
\midrule
PPO & Qwen2.5-1.5B-Instruct & 59.1 & \textbf{65.2}$_{{+6.1}}$ \\
RLOO & Qwen2.5-1.5B-Instruct & 56.7 & \textbf{61.2}$_{{+4.5}}$ \\
GRPO & Qwen3-1.7B & 45.4 & \textbf{54.2}$_{{+8.8}}$ \\
GRPO & Llama3.2-3B-Instruct & 4.0 & \textbf{62.2}$_{{+58.2}}$ \\
GRPO & Qwen2.5-14B-Instruct & 74.7 & \textbf{77.1}$_{{+2.4}}$ \\
\bottomrule
\end{tabular}
}
\vspace{-0.05in}
\caption{WebShop success rate (\%) across different RL algorithms and base models.}
\label{tab:multi_rl_model}
\end{table}

\subsection{Experimental Results}
Table~\ref{tab:main} shows that \methodname{} consistently improves both GRPO and GIGPO on ALFWorld and WebShop across model scales.
On ALFWorld, \methodname{} improves the overall success rate of GRPO by $+7.9$ and $+3.0$ for Qwen2.5-1.5B-Instruct and Qwen2.5-7B-Instruct, respectively.
It also improves GIGPO by $+2.8$ and $+1.0$.
On WebShop, \methodname{} yields even larger success-rate gains, improving GRPO by $+8.0$ and $+4.0$ and GIGPO by $+9.1$ and $+2.9$ at 1.5B and 7B, respectively.
These results indicate that rollout-based world-model co-training improves long-horizon agent decision-making without adding extra models or changing inference.

Table~\ref{tab:main_qa} further evaluates \methodname{} on multi-turn search-augmented QA tasks.
For GRPO, \methodname{} improves the average score from $43.9\%$ to $44.8\%$ at 3B and from $47.9\%$ to $48.8\%$ at 7B.
For GIGPO, the average score increases from $41.2\%$ to $44.2\%$ at 3B and from $45.8\%$ to $47.5\%$ at 7B.
The gains across interactive environments and search-augmented QA suggest that \methodname{} is complementary to different RL algorithms and generalizes across agentic tasks. More results can be found in \cref{apx:results}.

\begin{figure}[!t]
  \includegraphics[width=0.9\columnwidth]{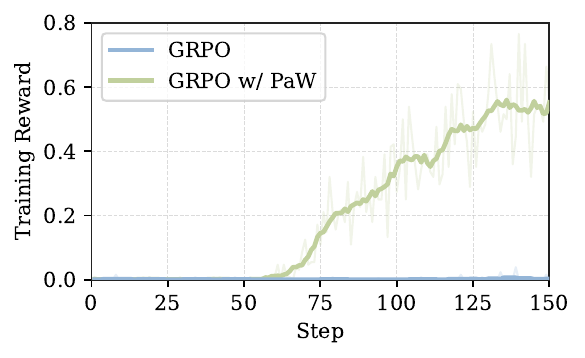}
  \vspace{-0.1in}
  \caption{Training rewards of Llama3.2-3B-Instruct on WebShop. Compared with vanilla GRPO, \methodname{} helps the agent escape sparse-reward failure and obtain positive success signals.}
  \label{fig:grpo_fail}
\end{figure}

\subsection{Different Models and RL algorithms}
\label{sec:diff_model_rl}

We further examine whether the gains from \methodname{} generalize beyond the main GRPO/GIGPO settings.
To test generality across RL algorithms, we combine \methodname{} with PPO~\cite{schulman2017proximal} and RLOO~\cite{kool2019buy,ahmadian2024back}.
To test generality across model backbones, we apply GRPO w/ \methodname{} to different model families and scales, including Qwen3-1.7B~\cite{qwen2025qwen3}, Llama3.2-3B-Instruct~\cite{Grattafiori2024Llama3}, and Qwen2.5-14B-Instruct.
Implementation details can be found in \cref{apx:implementation}.
As shown in \cref{tab:multi_rl_model}, \methodname{} consistently improves WebShop success rate across RL algorithms, model families, and model scales.
It improves PPO and RLOO by $+6.1$ and $+4.5$, respectively, showing that the proposed world-model co-training objective is not tied to a specific RL algorithm.
The improvement also persists across different backbones, including a $+2.4$ gain on Qwen2.5-14B-Instruct.
Notably, \methodname{} improves Llama3.2-3B-Instruct from $4.0\%$ to $62.2\%$, suggesting that world-model supervision can provide useful learning signals when vanilla RL struggles to obtain positive rewards; we analyze this case in \cref{subsec:work_on_hard}.

\subsection{\methodname{} Helps Where RL Fails}
\label{subsec:work_on_hard}
We next study whether \methodname{} can help in sparse-reward settings where vanilla RL struggles to learn from on-policy rollouts.
We train Llama3.2-3B-Instruct on WebShop with GRPO and compare the training dynamics with and without \methodname{}.
As shown in \cref{fig:grpo_fail}, vanilla GRPO rarely obtains positive rewards in this challenging setting, causing the training signal to collapse.
In contrast, \methodname{} provides dense WM supervision by predicting next observations from collected state-action transitions.
This auxiliary supervision allows the model to learn useful transition information even when most trajectories receive zero task reward.
After several training steps, the model begins to generate successful rollouts, which in turn provides meaningful RL signals for further policy improvement.
This result suggests that \methodname{} can improve RL robustness by mitigating sparse-reward failures, especially for weaker base models or harder tasks. This analysis also corresponds to the $+58.2$ improvement on Llama3.2-3B-Instruct reported in \cref{tab:multi_rl_model}.

\begin{figure}[t]
  \includegraphics[width=0.9\columnwidth]{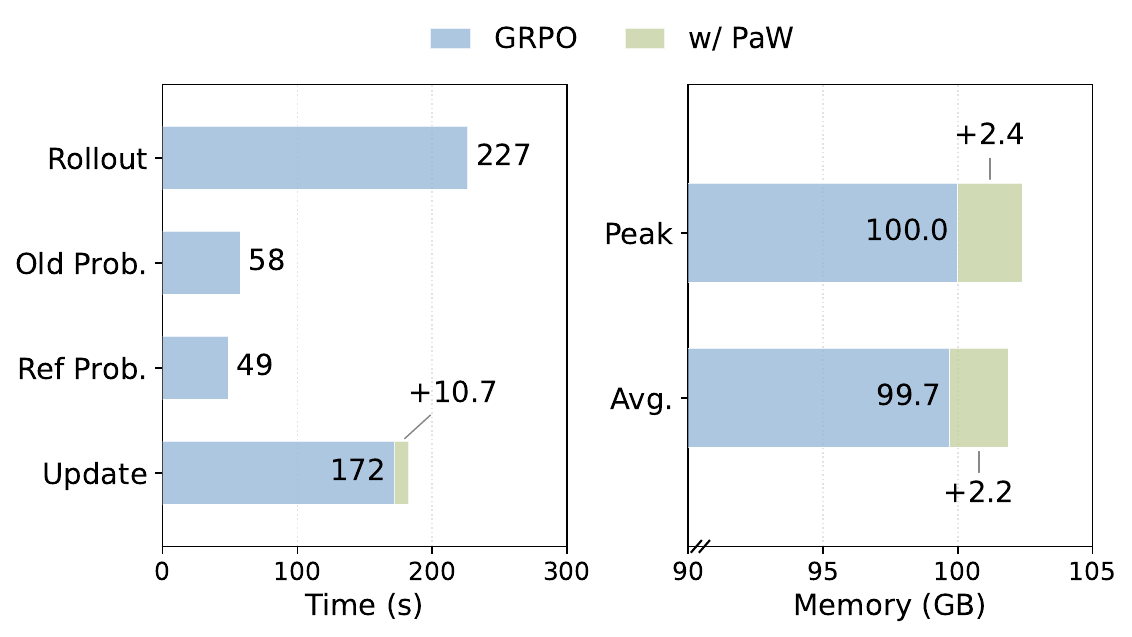}
  \vspace{-0.1in}
  \caption{Per-step training time and GPU memory breakdown on ALFWorld with Qwen2.5-1.5B-Instruct. \methodname{} increases both wall-clock time and GPU memory usage by approximately $2\%$.}
  \label{fig:overhead}
\end{figure}

\subsection{Computational Overhead}

We further measure the computational overhead introduced by \methodname{}.
We profile the per-step wall-clock time and GPU memory usage of vanilla GRPO and GRPO w/ \methodname{} on ALFWorld with Qwen2.5-1.5B-Instruct.
As shown in \cref{fig:overhead}, \methodname{} introduces negligible overhead: it reuses the same rollout data and actor forward pass, adding only the WM objective during the RL update while leaving the rest of the pipeline unchanged.
It adds only $10.7$\,s per step ($2.1\%$ of the $\sim\!505$\,s GRPO step time), with peak and average GPU memory increasing by $2.4$\,GB ($2.4\%$) and $2.2$\,GB ($2.2\%$), respectively.
Thus, \methodname{} improves agent performance with minimal training overhead.
A detailed analysis is provided in \cref{apx:overhead}.

\begin{table}
  \centering
  \resizebox{0.8\columnwidth}{!}{
\begin{tabular}{lcc}
\toprule
\textbf{Method} & \textbf{ALFWorld} & \textbf{WebShop} \\
\midrule
GRPO & 70.0 & 60.6 \\
\rowcolor{gray!25}\textbf{GRPO w/ \methodname{}} & \textbf{77.9} & \textbf{68.6} \\
~ w/o Ada. Coef. & 75.5 & 67.0 \\
~ w/ CE loss & 68.5 & 57.2 \\
\bottomrule
\end{tabular}
}
\vspace{-0.05in}
  \caption{Ablation results on ALFWorld and WebShop with Qwen2.5-1.5B-Instruct using GRPO as the base RL algorithm. ``w/o Ada. Coef.'' sets $\lambda_{\mathrm{WM},g}=1$ for all rollout groups, and ``w/ CE loss'' replaces the CMAE observation loss with standard cross-entropy.}
  \label{tab:ablation}
\end{table}

\subsection{Ablation Study}
\cref{tab:ablation} studies the reward-adaptive WM coefficient and the clipped MAE WM loss in \methodname{} on ALFWorld and WebShop with Qwen2.5-1.5B-Instruct and GRPO .
Fixing $\lambda_{\mathrm{WM},g}=1$ still outperforms vanilla GRPO but reduces performance from $77.9\%$ to $75.5\%$ on ALFWorld and from $68.6\%$ to $67.0\%$ on WebShop, showing the benefit of adaptive loss balancing.
Replacing CMAE with standard cross-entropy causes a larger drop, to $68.5\%$ on ALFWorld and $57.2\%$ on WebShop, indicating that CE can overfit noisy observation tokens while CMAE provides more robust supervision.
Together, both components are important for effective world modeling co-training.

\subsection{Valid Transitions Drive the Gains}

We further examine whether the gains of \methodname{} arise from learning valid action-conditioned transitions rather than from generic dense gradients.
We compare two controls under the same Qwen2.5-1.5B-Instruct and GRPO setting: \emph{AOShuffled} randomly permutes action--observation pairs, while \emph{Random} replaces observation targets with random tokens.
As shown in \cref{tab:transition_controls}, both controls perform close to vanilla GRPO and substantially worse than \methodname{}.
The small improvement from AOShuffled may reflect auxiliary language-modeling regularization, while the full gain requires preserving the correct action--observation correspondence.

\begin{table}[h]
\centering
\small
\setlength{\tabcolsep}{10pt}
\renewcommand{\arraystretch}{1.08}
\begin{tabular}{@{}lcc@{}}
\toprule
\textbf{Method} & \textbf{ALFWorld} & \textbf{WebShop} \\
\midrule
GRPO                                  & 70.0          & 60.6          \\
\rowcolor{gray!20}
\quad w/ \methodname{}               & \textbf{77.9} & \textbf{68.6} \\
\quad w/ \methodname{} (AOShuffled)  & 72.2          & 61.2          \\
\quad w/ \methodname{} (Random)      & 69.8          & 60.8          \\
\bottomrule
\end{tabular}
\vspace{-0.05in}
\caption{Success rates (\%) under valid and corrupted transition supervision.}
\label{tab:transition_controls}
\end{table}

\subsection{Generalization Performance}
We further evaluate whether \methodname{} generalizes beyond the environments observed during training.
We train Qwen2.5-1.5B-Instruct on the ALFWorld seen split and evaluate it on unseen ALFWorld environments and on WebShop to assess cross-domain generalization.
As shown in \cref{tab:cross_task_generalization}, \methodname{} outperforms GRPO on both ALFWorld-Unseen ($66.7\% \rightarrow 76.3\%$) and WebShop ($5.7\% \rightarrow 10.8\%$), suggesting that the learned transition knowledge generalizes across environment instances and task domains.

\begin{table}[t]
\centering
\resizebox{\columnwidth}{!}{
\begin{tabular}{lccc}
\toprule
\multirow{2}{*}{\textbf{Method}} & \textbf{ALFWorld-Unseen} & \multicolumn{2}{c}{\textbf{WebShop}} \\
& Succ. & Score & Succ. \\
\midrule
GRPO & 66.7 & 13.2 & 5.7 \\
GRPO w/ \methodname{} & \textbf{76.3} & \textbf{44.8} & \textbf{10.8} \\
\bottomrule
\end{tabular}
}
\vspace{-0.05in}
\caption{Generalization performance of Qwen2.5-1.5B-Instruct trained on the ALFWorld seen split.}
\label{tab:cross_task_generalization}
\end{table}

\subsection{Hyperparameter Sensitivity}
\begin{figure}[hbt]
  \includegraphics[width=\columnwidth]{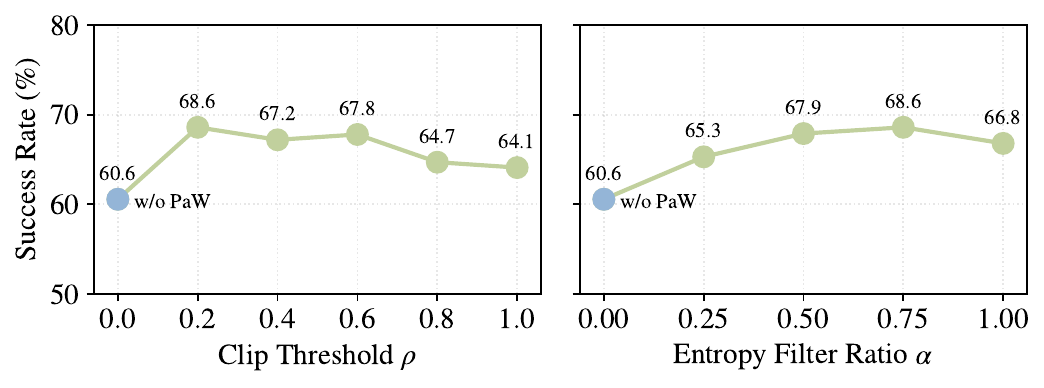}
  \vspace{-0.3in}
  \caption{Hyperparameter sensitivity on WebShop with Qwen2.5-1.5B-Instruct using GRPO as the base RL algorithm. We vary the CMAE clipping threshold $\rho$ and the entropy selection ratio $\alpha$.}
  \label{fig:hyper_analysis}
\end{figure}

We further analyze the sensitivity of \methodname{} to the entropy selection ratio $\alpha$ and clipping threshold $\rho$ on WebShop with Qwen2.5-1.5B-Instruct and GRPO.
As shown in \cref{fig:hyper_analysis}, \methodname{} remains effective across a broad range of values.
Moderate clipping works best, with $\rho=0.2$ performing the strongest, while overly large thresholds reduce performance, highlighting the importance of filtering unpredictable observation tokens.
Performance also varies smoothly with $\alpha$, with $\alpha=0.75$ giving the best result, suggesting that entropy-based transition selection is useful but not overly sensitive to the exact ratio.
Overall, \methodname{} does not require delicate hyperparameter tuning.

%% file: section/related_work.tex
\section{Related Work}

\paragraph{Training LLM agents.}
LLM agents map instructions, interaction histories, and observations into executable actions for web, tool-use, embodied, and other interactive tasks~\cite{wang2023survey,yao2023react,shinn2024reflexion}.
Beyond prompting, recent work trains agents with supervised fine-tuning or reinforcement learning~\cite{deng2023mind2web,zeng2023agenttuning,chen2023fireact,xi2024agentgym,jin2025search,sun2025zerosearch,feng2025group}.
Because agent rewards are often sparse and delayed, existing RL methods mainly improve credit assignment or add auxiliary training signals~\cite{feng2025group,fang2026proxmo,lu2026sdar}.
\methodname{} is complementary to these approaches: instead of changing the policy-gradient estimator, it reuses the same on-policy rollouts to provide dense next-observation supervision.

\paragraph{World modeling for language agents.}
World models learn environment dynamics by predicting future states or rewards~\cite{ha2018worldmodels,hafner2020dreamer,hafner2021mastering}.
Recent language-agent methods use LLMs as world models, simulators, or transition predictors for planning, verification, and policy learning~\cite{lin2023dynalang,guo2025worldmodelling,chae2025webagentsworldmodels,yu2026rwml,liu2026imaginethenplan}.
While these methods demonstrate the value of future-observation modeling, they often require a separate world model, additional simulator training, or inference-time planning.
In contrast, \methodname{} folds world modeling into standard on-policy RL by training the same policy model to predict next observations from its own rollouts, introducing no extra interactions, no separate model, and no additional inference-time computation. Additional related work is discussed in \cref{apx:related_work}.

%% file: section/conclusion.tex
\section{Conclusion}
In this paper, we presented \methodname{}, a policy and world modeling co-training framework for LLM agents.
Rather than training a separate simulator or adding inference-time planning, \methodname{} reuses on-policy RL rollouts as action-conditioned next-observation supervision and optimizes an auxiliary world-modeling loss on the same policy model.
To make this supervision effective in noisy agentic environments, \methodname{} combines action-entropy-based transition selection, clipped MAE observation prediction, and reward-adaptive loss balancing.
Experiments on ALFWorld, WebShop, and search-augmented QA show consistent gains across RL algorithms and model scales, with minimal training overhead and no additional inference cost.
These results suggest that standard RL rollouts provide a simple and practical source of world modeling supervision for improving language-agent training.

\section{Limitations}

While \methodname{} consistently improves RL performance, there exist two main limitations.
First, it relies on one-step next-observation supervision, which captures local dynamics but does not explicitly model longer-horizon dependencies or compounding prediction errors.
Extending the co-training objective to multi-step world modeling is a promising direction for future work.
Second, our experiments focus on relatively simple agent tasks such as web browsing. More complex scenarios, such as software engineering tasks, are left for future exploration.

\section*{Acknowledgments}
This work was supported by National Key Research and Development Program of China under Grant 2022YFA1004102,
and in part by the National Natural Science Foundation of China under Grants T2495254, 62502192.

%% file: appendix/other_implementation.tex
\clearpage

\section{Implementation Details}
\label{apx:implementation}

\subsection{Details of Training}\label{appendix:train_detail}
\paragraph{Hyperparameters for ALFWorld.}
All methods are configured with identical hyperparameters: the maximum prompt length is 2048 tokens, and the maximum response length is 512 tokens. Each episode allows up to 50 environment steps. The learning rate is set to 1e-6 for the actor and 1e-5 for the critic (used only in PPO). We adopt a rule-based reward, assigning a reward of 10 for success and 0 for failure ($R_{\text{max}} = 10$). To handle invalid actions generated by the agent, we apply a reward penalty of -0.1. For all group-based RL methods, we use a group size of 8 and sample 16 different groups per rollout, resulting in a total of 128 environments.
In contrast, PPO uses 128 separate environments for rollouts. 
For GIGPO, the weighting coefficient $\omega$ is fixed at 1 without further tuning, and the discount factor $\gamma$ is set to 0.95.
And we use the normalized version of GIGPO.
The rollout temperature is set to 1.0, while the validation temperature is set to 0.4. The mini-batch size is 256 
The history number is set to 2.

\paragraph{Hyperparameters for WebShop.}
All methods are configured with identical hyperparameters: the maximum prompt length is 4096 tokens, and the maximum response length is 512 tokens. Each episode is limited to 15 environment steps. The learning rate is 1e-6 for the actor and 1e-5 for the critic (used only in PPO). We adopt a rule-based reward, assigning a reward of 10 for success and 0 for failure. So the $R_{\text{max}} = 10$. Invalid actions are penalized with a reward of -0.1. As with ALFWorld, all group-based RL methods use a group size of 8 and sample 16 groups per rollout, totaling $16 \times 8 = 128$ environments. PPO, on the other hand, uses 128 distinct environments for rollouts. 
For GIGPO, the weighting coefficient $\omega$ is set to 1 without additional tuning, and the discount factor $\gamma$ is set to 0.95.
And we use the normalized version of GIGPO.
The rollout temperature is set to 1.0, while the validation temperature is set to 0.4. The mini-batch size is 64. 

\paragraph{Hyperparameters for Search-Augmented QA.}
The maximum prompt length is 4096 tokens, and the maximum response length is 512 tokens. The max turn is set to 4. The learning rate is 1e-6 for the actor. We adopt a rule-based reward, assigning a reward of 1 for success and 0 for failure. So the $R_{\text{max}} = 1$. Invalid actions are penalized with a reward of -0.01. We set the train data size to 256 and use a group size of 5. 
For GIGPO, the weighting coefficient $\omega$ is set to 1 without additional tuning, the discount factor $\gamma$ is set to 0.95, and the similarity threshold is set to 0.9.
Rollout and validation temperatures are set to 1.0 and 0.0, respectively. The mini-batch size is 512.


\begin{figure}[t]
\centering
\resizebox{\columnwidth}{!}{
\begin{tcolorbox}[colback=gray!5!white, colframe=black!75!black, 
title=Prompt Template for ALFWorld, boxrule=0.3mm, width=\columnwidth, arc=3mm, auto outer arc=true]
You are an expert agent operating in the ALFRED embodied Environment. Your task is to: \textcolor{brown}{\{task\_description\}}. Prior to this step, you have already taken \textcolor{brown}{\{step\_count\}} step(s). Below are the most recent \textcolor{brown}{\{history\_length\}} observations and the corresponding actions you took: \textcolor{brown}{\{action\_history\}}. You are now at step \textcolor{brown}{\{current\_step\}} and your current observation is: \textcolor{brown}{\{current\_observation\}}. Your admissible actions of the current situation are: [\textcolor{brown}{\{admissible\_actions\}}].

Now it's your turn to take an action.
You should first reason step-by-step about the current situation. This reasoning process MUST be enclosed within \textcolor{deepgreen}{<think>} \textcolor{deepgreen}{</think>} tags.
Once you've finished your reasoning, you should choose an admissible action for current step and present it within \textcolor{deeppurple}{<action>} \textcolor{deeppurple}{</action>} tags.
\end{tcolorbox}
}
\vspace{-0.2in}
\caption{The prompt template of ALFWorld agents.}
\label{prompt:alfworld_prompt_temp}
\end{figure}

\begin{figure}[t]
\centering
\resizebox{\columnwidth}{!}{
\begin{tcolorbox}[colback=gray!5!white, colframe=black!75!black, 
title=Prompt Template for WebShop, boxrule=0.3mm, width=\columnwidth, arc=3mm, auto outer arc=true]
You are an expert autonomous agent operating in the WebShop e‑commerce environment. Your task is to: \textcolor{brown}{\{task\_description\}}. Prior to this step, you have already taken \textcolor{brown}{\{step\_count\}} step(s). Below are the most recent \textcolor{brown}{\{history\_length\}} observations and the corresponding actions you took: \textcolor{brown}{\{action\_history\}}. You are now at step \textcolor{brown}{\{current\_step\}} and your current observation is: \textcolor{brown}{\{current\_observation\}}. Your admissible actions for the current situation are: [\textcolor{brown}{\{available\_actions\}}].

Now it's your turn to take one action for the current step.
You should first reason step-by-step about the current situation, then think carefully which admissible action best advances the shopping goal. This reasoning process MUST be enclosed within \textcolor{deepgreen}{<think>} \textcolor{deepgreen}{</think>} tags.
Once you've finished your reasoning, you should choose an admissible action for current step and present it within \textcolor{deeppurple}{<action>} \textcolor{deeppurple}{</action>} tags.
\end{tcolorbox}
}
\vspace{-0.2in}
\caption{The prompt template of WebShop agents.}
\label{prompt:webshop_prompt_temp}
\end{figure}

\begin{figure}[t]
\centering
\resizebox{\columnwidth}{!}{
\begin{tcolorbox}[colback=gray!5!white, colframe=black!75!black, 
title=Prompt Template for Search, boxrule=0.3mm, width=\columnwidth, arc=3mm, auto outer arc=true]
You are an expert agent tasked with answering the given question step-by-step. Your question: \textcolor{brown}{\{task\_description\}}. Prior to this step, you have already taken \textcolor{brown}{\{step\_count\}} step(s). Below is the interaction history where \textcolor{deeppurple}{<search>} \textcolor{deeppurple}{</search>} wrapped your past search queries and \textcolor{deepblue}{<information>} \textcolor{deepblue}{</information>} wrapped the corresponding search results returned by the external search engine. History: \textcolor{brown}{\{memory\_context\}}

Now it's your turn to respond for the current step.
You should first conduct reasoning process. This process MUST be enclosed within \textcolor{deepgreen}{<think>} \textcolor{deepgreen}{</think>} tags.
After completing your reasoning, choose only one of the following actions (do not perform both):

(1) If you find you lack some knowledge, you can call a search engine to get more external information using format: \textcolor{deeppurple}{<search>} your query \textcolor{deeppurple}{</search>}.

(2) If you have enough knowledge to answer the question confidently, provide your final answer within \textcolor{deeppurple}{<answer>} \textcolor{deeppurple}{</answer>} tags, without detailed illustrations. For example, \textcolor{deeppurple}{<answer>}Beijing\textcolor{deeppurple}{</answer>}.
\end{tcolorbox}
}
\vspace{-0.2in}
\caption{The prompt template of Search agents.}
\label{prompt:search_prompt_temp}
\end{figure}

\subsection{Baseline Details}
\label{appendix:comparing}

\begin{itemize}[leftmargin=*,nosep]
\item \emph{GPT-4o:} A closed-source, large-scale LLM used as a baseline for multi-turn agentic tasks~\citep{achiam2023gpt}.
\item \emph{Gemini-2.5-Pro:} Another closed-source LLM, comparable in scale and capability to GPT-4o~\citep{team2023gemini}.
\item \emph{ReAct:} A prompting-based agent that integrates reasoning and acting in an interleaved chain-of-thought framework \cite{yao2023react}.
\item \emph{Reflexion:} A prompting agent that incorporates self-reflection and iterative improvement over generated outputs~\citep{shinn2024reflexion}.
\item \emph{PPO:} Proximal Policy Optimization, a classic RL algorithm for policy learning~\citep{schulman2017proximal}.
\item \emph{RLOO:} Reinforcement Learning with Offline Observations, a group-based RL approach that estimates advantages without value networks~\citep{kool2019buy,ahmadian2024back}.
\item \emph{GRPO:} Group-based RL with trajectory-level advantage estimation, designed to scale RL to multi-step tasks~\citep{shao2024deepseekmath}.
\item \emph{GiGPO:} Grouped Incremental GPO, a prior hierarchical RL method that performs group-wise advantage estimation for LLM-based agents~\citep{feng2025group}.
\end{itemize}

\subsection{Prompts}


The prompt templates used for the LLM agents are shown in \cref{prompt:alfworld_prompt_temp}, \cref{prompt:webshop_prompt_temp}, and \cref{prompt:search_prompt_temp}. Each template is implemented using Python-style string formatting, where fields enclosed in curly braces (\textcolor{brown}{\{\}}) denote semantic slots that are instantiated at runtime via Python's \texttt{.format()} function. For example, placeholders such as \textcolor{brown}{\{task\_description\}}, \textcolor{brown}{\{step\_count\}}, and \textcolor{brown}{\{current\_observation\}} are dynamically replaced with task-specific context. To provide the agent with temporal context, we additionally incorporate interaction history: we retain the two most recent history steps for ALFWorld and WebShop, and use the complete history for search-augmented question answering.

\begin{table*}[t]
\centering
\resizebox{\textwidth}{!}{
\begin{tabular}{llccccccc|cc}
\toprule
\multirow{2}{*}{Type} & \multirow{2}{*}{Method} & \multicolumn{7}{c|}{\textbf{ALFWorld}} & \multicolumn{2}{c}{\textbf{WebShop}} \\
 & & Pick & Look & Clean & Heat & Cool & Pick2 & All & Score & Succ.\\
\midrule
\multicolumn{11}{l}{\textit{Qwen2.5-1.5B-Instruct}} \\
RL Training& GRPO & 86.5\textsubscript{\textpm4.3} & 46.3\textsubscript{\textpm8.4} & 79.0\textsubscript{\textpm3.2} & 70.2\textsubscript{\textpm2.5} & 69.1\textsubscript{\textpm8.0} & 47.8\textsubscript{\textpm5.8} & 70.0\textsubscript{\textpm4.2} & 75.6\textsubscript{\textpm3.8} & 60.6\textsubscript{\textpm4.6} \\
\rowcolor{gray!25}RL Training& \textbf{GRPO w/ \methodname{}} & \textbf{87.8}\textsubscript{\textpm3.6} & \textbf{59.3}\textsubscript{\textpm8.9} & \textbf{84.5}\textsubscript{\textpm1.7} & \textbf{73.7}\textsubscript{\textpm4.3} & \textbf{75.4}\textsubscript{\textpm8.4} & \textbf{69.6}\textsubscript{\textpm3.5} & \textbf{77.9}\textsubscript{\textpm2.7} & \textbf{83.8}\textsubscript{\textpm2.1} & \textbf{68.6}\textsubscript{\textpm2.9} \\
RL Training& GIGPO & 95.3\textsubscript{\textpm2.2} & 84.3\textsubscript{\textpm4.6} & 87.7\textsubscript{\textpm1.2} & 92.6\textsubscript{\textpm7.0} & 79.8\textsubscript{\textpm1.4} & 82.3\textsubscript{\textpm6.1} & 87.6\textsubscript{\textpm1.6} & 83.2\textsubscript{\textpm1.9} & 66.2\textsubscript{\textpm3.4} \\
\rowcolor{gray!25}RL Training& \textbf{GIGPO w/ \methodname{}} & \textbf{95.3}\textsubscript{\textpm4.5} & 83.3\textsubscript{\textpm4.5} & \textbf{91.8}\textsubscript{\textpm6.0} & 89.5\textsubscript{\textpm4.3} & \textbf{89.1}\textsubscript{\textpm0.3} & \textbf{84.5}\textsubscript{\textpm5.8} & \textbf{90.4}\textsubscript{\textpm1.3} & \textbf{87.7}\textsubscript{\textpm1.7} & \textbf{75.3}\textsubscript{\textpm3.2} \\
\midrule
\multicolumn{11}{l}{\textit{Qwen2.5-7B-Instruct}} \\
RL Training& GRPO & 90.8\textsubscript{\textpm5.1} & 66.1\textsubscript{\textpm6.7} & 89.3\textsubscript{\textpm5.4} & 74.7\textsubscript{\textpm6.9} & 72.5\textsubscript{\textpm5.4} & 64.7\textsubscript{\textpm7.3} & 77.6\textsubscript{\textpm5.2} & 75.4\textsubscript{\textpm2.7} & 66.5\textsubscript{\textpm3.6} \\
\rowcolor{gray!25}RL Training& \textbf{GRPO w/ \methodname{}} & 90.4\textsubscript{\textpm6.8} & \textbf{80.7}\textsubscript{\textpm7.4} & 86.8\textsubscript{\textpm6.8} & \textbf{82.9}\textsubscript{\textpm8.6} & \textbf{76.5}\textsubscript{\textpm2.6} & \textbf{67.3}\textsubscript{\textpm3.7} & \textbf{80.6}\textsubscript{\textpm3.1} & \textbf{84.5}\textsubscript{\textpm2.3} & \textbf{70.5}\textsubscript{\textpm2.9} \\
RL Training& GIGPO & 97.7\textsubscript{\textpm1.6} & 82.7\textsubscript{\textpm7.9} & 98.8\textsubscript{\textpm1.6} & 83.7\textsubscript{\textpm7.2} & 89.3\textsubscript{\textpm8.2} & 79.2\textsubscript{\textpm6.6} & 90.8\textsubscript{\textpm1.3} & 85.0\textsubscript{\textpm2.9} & 73.8\textsubscript{\textpm3.2} \\
\rowcolor{gray!25}RL Training& \textbf{GIGPO w/ \methodname{}} & \textbf{98.2}\textsubscript{\textpm1.0} & \textbf{85.6}\textsubscript{\textpm6.2} & 98.6\textsubscript{\textpm1.1} & \textbf{84.5}\textsubscript{\textpm7.6} & \textbf{91.5}\textsubscript{\textpm7.2} & \textbf{84.3}\textsubscript{\textpm6.8} & \textbf{91.8}\textsubscript{\textpm1.2} & \textbf{87.6}\textsubscript{\textpm2.4} & \textbf{76.7}\textsubscript{\textpm2.7} \\
\bottomrule
\end{tabular}
}
\vspace{-0.1in}
\caption{Full performance on ALFWorld and WebShop with statistical stats. Results are averaged over 3 random seeds. For ALFWorld, we report the average success rate (\%) for each subtask as well as the overall result. For WebShop, we report both the average score and the average success rate (\%).}
\label{tab:main_std_full}
\end{table*}

The \textcolor{deepgreen}{<think>} \textcolor{deepgreen}{</think>} block is used to elicit explicit step-by-step reasoning from the agent, encouraging chain-of-thought~\cite{wei2022chain} style deliberation. The \textcolor{deeppurple}{<action>} \textcolor{deeppurple}{</action>} block specifies the agent's final action decision. For the search agent, reasoning traces are produced within \textcolor{deepgreen}{<think>} \textcolor{deepgreen}{</think>} tags, search queries are issued within \textcolor{deeppurple}{<search>} \textcolor{deeppurple}{</search>} tags, and final answers are provided within \textcolor{deeppurple}{<answer>} \textcolor{deeppurple}{</answer>} tags. Retrieved evidence from the retriever is supplied to the agent using \textcolor{deepblue}{<information>} \textcolor{deepblue}{</information>} tags.

\subsection{Noise-Gradient Analysis}
\label{apx:noise_gradient_analysis}

This section describes the diagnostic used for panel (c) of \cref{fig:cmae-visualization}.
The goal is to measure how much of the world-model gradient budget is assigned to unpredictable WebShop observation tokens under CE and MAE losses.
We run the diagnostic on WebShop with Qwen2.5-1.5B-Instruct.
We randomly sample 100 WebShop search-result steps from rollout traces.
For each sampled transition $(\vh_t,\va_t,\vo_{t+1})$, we use the same serialization format as world-model training.
The decision context $\vh_t$ and the generated action $\va_t$ form the prefix, and the next observation $\vo_{t+1}$ is treated as the teacher-forced target.
We only score tokens that belong to the next-observation span.

For the $i$-th target token in $\vo_{t+1}$, we compute the model probability $p_{t,i}:=\vpitheta\!\left(\vo_{t+1}^{(i)}\mid\vh_t,\va_t,\vo_{t+1}^{(<i)}\right)$,
following the notation of \cref{sec:method:cmae}.
We then compare two token-level losses.
The CE diagnostic uses $\ell_{\mathrm{CE}}^{(i)}=-\log p_{t,i}$, matching the unbounded WM CE objective in \cref{eq:wm-ce}.
The MAE diagnostic uses $\ell_{\mathrm{MAE}}^{(i)}=1-p_{t,i}$, matching the bounded per-token term used by our Clipped MAE loss in \cref{eq:cmae-transition-loss} before applying the confidence mask $m_{t,i}$.
To compare gradient allocation rather than raw loss scale, we compute the target-logit gradient magnitude for each token, consistent with the parameter-gradient comparison in \cref{eq:ce-mae-gradient}.
For CE this magnitude is $1-p_{t,i}$.
For MAE this magnitude is $p_{t,i}(1-p_{t,i})$.
We sum these magnitudes within each token category and normalize them by the total gradient magnitude over all scored next-observation tokens.

We define noisy tokens using human-written WebShop rules.
A token is labeled as noisy if its character span overlaps either a random product identifier or a brand string.
Random product identifiers are matched with ASIN-style pattern \texttt{B[0-9A-Z]\{9\}}.
Brand strings are matched as brand-like uppercase spans with at least four letters, optionally followed by digits.
This threshold excludes structural markers such as \texttt{SEP}.
All other next-observation tokens are labeled as meaningful for this diagnostic.
When a word is split into multiple subword tokens, each subword inherits  noisy label if its offset overlaps a noisy character span.

This analysis is evaluation-only and does not update model parameters.
It isolates the effect of the loss function on gradient allocation over the same sampled tokens.
The resulting normalized shares show whether a loss over-allocates gradient to noisy WebShop surface strings.

\subsection{Computational Overhead of \methodname{}}
\label{apx:overhead}

\paragraph{Overall overhead.}
\methodname{} introduces no additional trainable parameters and increases training time and memory usage by only approximately $2\%$, as reported in \cref{fig:overhead}.

\paragraph{Stage-wise analysis.}
We decompose the training pipeline to identify where \methodname{} incurs additional computation relative to vanilla on-policy RL.
\begin{itemize}
    \item \textbf{Rollout (no overhead).}
    Both methods use the same standard generation procedure.

    \item \textbf{Old log-probability computation (no overhead).}
    Both methods perform the same computation required by the base RL objective.

    \item \textbf{Policy update (additional overhead).}
    Vanilla RL computes the loss over action tokens, whereas \methodname{} additionally processes observation tokens and backpropagates the world-modeling loss. This is the only source of additional computation.

    \item \textbf{KL regularization (no overhead).}
    Both methods use the same current- and reference-policy computations.
\end{itemize}

%% file: appendix/other_main_results.tex
\begin{figure*}[t]
  \centering
  \begin{tabular}{ccc}
    \includegraphics[width=0.31\textwidth]{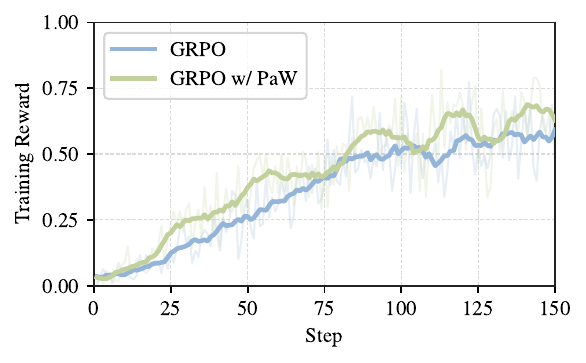} &
    \includegraphics[width=0.31\textwidth]{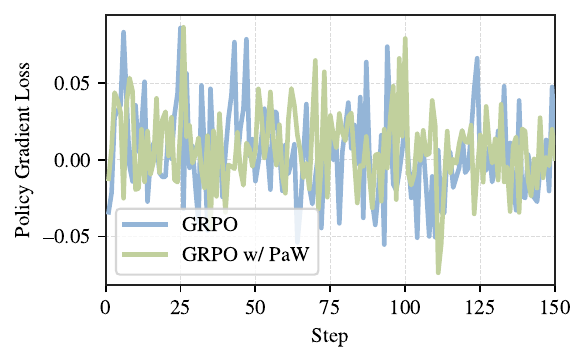} &
    \includegraphics[width=0.31\textwidth]{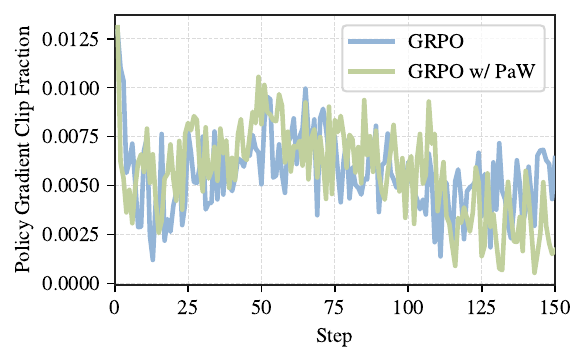} \\
    \small (a) Training reward. &
    \small (b) Policy-gradient loss. &
    \small (c) Clipped update ratio.
  \end{tabular}
  \vspace{-0.1in}
  \caption{
    Policy-side training dynamics on WebShop.
    \methodname{} improves the training reward over the GRPO baseline, while the policy-gradient loss and clipped update ratio remain broadly comparable.
    This suggests that the auxiliary world modeling objective improves learning without substantially changing the main policy-optimization dynamics.
  }
  \label{fig:apx:policy_training_dynamics}
\end{figure*}

\begin{figure*}[!t]
  \centering
  \begin{tabular}{ccc}
    \includegraphics[width=0.31\textwidth]{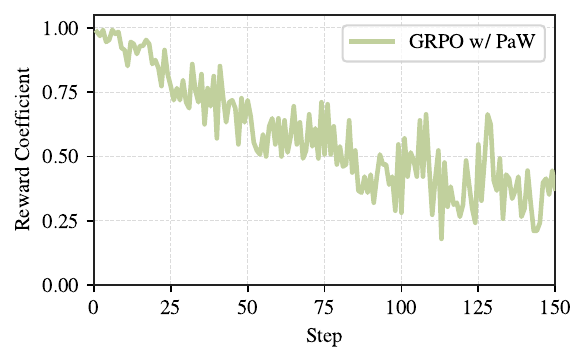} &
    \includegraphics[width=0.31\textwidth]{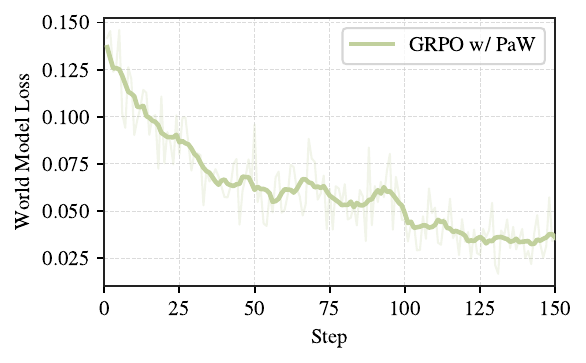} &
    \includegraphics[width=0.31\textwidth]{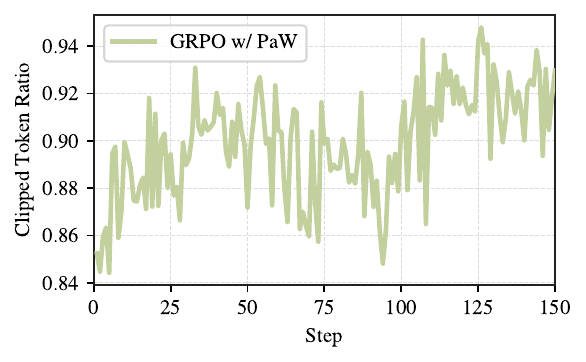} \\
    \small (a) Adaptive coefficient. &
    \small (b) World modeling loss. &
    \small (c) Clipped token ratio.
  \end{tabular}
  \vspace{-0.1in}
  \caption{
    \methodname{}-side training dynamics.
    The reward-adaptive coefficient decreases as training reward improves, the world modeling loss trends downward, and the clipped token ratio reports how often the margin objective clips observation-token supervision.
  }
  \label{fig:apx:paw_training_dynamics}
\end{figure*}

\begin{table*}[tb]
\centering
\resizebox{\textwidth}{!}{
\begin{tabular}{lllc}
\toprule
Category & Token & Context & Probability \\
\midrule
\multirow{3}{*}{Lowest probability}
& \texttt{JR}  & \texttt{[SEP] 'B07DKG}\evaltok{JR}\texttt{74' [SEP] 'Amazon Brand - Daily Ritual'} & $9.4\times10^{-13}$ \\
& \texttt{CBD} & \texttt{[SEP] 'B00IT}\evaltok{CBD}\texttt{3Y' [SEP] 'Laredo Womens Spellbound Studded'} & $7.0\times10^{-12}$ \\
& \texttt{HR}  & \texttt{[SEP] 'B09Q5Z}\evaltok{HR}\texttt{VM' [SEP] 'Yinimo Mens Gym Shorts Fashionable'} & $1.0\times10^{-11}$ \\
\midrule
\multirow{3}{*}{Near threshold}
& \texttt{2} & \texttt{Happy Birthday Mickey Confetti T-Shirt [SEP] '\$2}\evaltok{2}\texttt{.99'} & $0.2$ \\
& \texttt{0} & \texttt{Fit V-Neck Pullover Sweater [SEP] '\$2}\evaltok{0}\texttt{.66 to \$29.2' [SEP]} & $0.2$ \\
& \texttt{2} & \texttt{Raglan Baseball Tee [SEP] '\$}\evaltok{2}\texttt{3.99' [SEP]} & $0.2$ \\
\bottomrule
\end{tabular}
}
\vspace{-0.05in}
\caption{Representative low-probability and near-threshold WebShop observation tokens. Square brackets mark the target token within its context.}
\label{tab:cmae_token_examples}
\end{table*}

\section{Additional Experimental Results}
\label{apx:results}

\subsection{Full Results of ALFWorld and WebShop}
\label{apx:full_results_alfworld_webshop}

\cref{tab:main_std_full} shows the full results on ALFWorld and WebShop, including the statistical stats. Results are averaged over 3 random seeds.
It shows that \methodname{} consistently improves the performance over the base RL methods.

\subsection{Training Dynamics}
\label{apx:training_dynamics}

\cref{fig:apx:policy_training_dynamics} compares the policy-side dynamics.
GRPO w/ \methodname{} obtains better training reward than the GRPO baseline, indicating that the additional world-model supervision improves the learned policy during RL.
At the same time, the policy-gradient loss and clipped update ratio are close to those of GRPO, suggesting that the improvement does not come from a large change in PPO-style update magnitude or clipping behavior.

\cref{fig:apx:paw_training_dynamics} further analyzes the auxiliary world-model objective.
The adaptive coefficient decreases during training, because higher-reward rollouts receive less world-model supervision under the reward-adaptive scaling rule.
Meanwhile, the world-model loss decreases, showing that the shared policy model learns the next-observation prediction target from RL rollouts.
The clipped token ratio tracks how many observation tokens are affected by the margin clipping mechanism, which keeps the auxiliary signal bounded as the world-model prediction improves.

\subsection{Analysis of Observation Tokens}
\label{apx:cmae_token_analysis}

We analyze which observation tokens receive reduced supervision under the clipped MAE objective.
Using the step-150 checkpoint of Qwen2.5-1.5B-Instruct, we collect token-level WM probabilities from 100 WebShop transitions, containing approximately 37K observation tokens.
Among 1,000 sampled instances with the lowest target-token probabilities, $81\%$ correspond to noisy surface details, primarily product identifiers, while only $0.2\%$ contain task-critical information according to human evaluation.
Among 500 token instances immediately above the clipping threshold, $79\%$ are identifiers, numbers, or prices.
These statistics indicate that clipped MAE primarily reduces optimization pressure on unpredictable surface details rather than decision-relevant state information.
Representative examples are shown in \cref{tab:cmae_token_examples}.

\subsection{\methodname{} Learns WM Knowledge}
\label{apx:wm_learn}

We further evaluate whether the model acquires action-conditioned transition knowledge under clipped MAE in \methodname{}.
Given an interaction history and the current action, we evaluate perplexity on the ground-truth next-observation tokens (PPL; lower is better) and accuracy on an eight-way multiple-choice prediction task (MCP Acc.; higher is better). MCP distractors are generated by executing alternative actions from the same state. We evaluate Qwen2.5-1.5B-Instruct on 256 unseen ALFWorld cases during training.

As shown in \cref{tab:wm_prediction}, GRPO degrades both metrics relative to the base model, whereas GRPO w/ \methodname{} reduces PPL to $1.38$ and improves MCP accuracy to $0.42$.
This confirms that the model learns useful WM knowledge under the clipped MAE WM loss.

\begin{table}[t]
\centering
\begin{tabular}{lcc}
\toprule
Model & PPL $\downarrow$ & MCP Acc. $\uparrow$ \\
\midrule
Base & 1.81 & 0.28 \\
GRPO & 1.98 & 0.22 \\
\rowcolor{gray!25}\textbf{GRPO w/ \methodname{}} & \textbf{1.38} & \textbf{0.42} \\
\bottomrule
\end{tabular}
\vspace{-0.05in}
\caption{Next-observation prediction on 256 unseen ALFWorld cases.}
\label{tab:wm_prediction}
\end{table}

%% file: appendix/other_related_work.tex
\section{Additional Related Work}
\label{apx:related_work}

\paragraph{LLM-based agents.}
Large language models have been widely used as autonomous agents that translate natural-language instructions, interaction histories, and environment observations into executable actions~\cite{wang2023survey,yao2023react,shinn2024reflexion}.
Such agents have been applied to a broad range of interactive domains, including software engineering~\cite{jimenez2024swebench,yang2024sweagent}, browser and web navigation~\cite{zhou2024webarena,qin2025uitars,deng2023mind2web}, algorithm design~\cite{lv2026ahdagent,zhang2026llmdriven}, safety~\cite{jiang-tang-2026-agents,LuLHOW024,lu2025safe}, science~\cite{Wei2025FromAF,hu2026agentmental,hu2025beyond}, and system control~\cite{rawles2025androidworld, li2025autogui, li2025uipro}.
Early agent systems mainly rely on prompting, tool use, memory, planning, or reflection to elicit multi-step behavior from frozen LLMs~\cite{yao2023react,shinn2024reflexion}.
Although these approaches can produce strong zero-shot or few-shot behaviors, their performance is often limited by distribution shift, long-horizon error accumulation, and the lack of task-specific adaptation.

\paragraph{Training language agents.}
To improve agent behavior beyond prompting, recent work trains LLM agents with supervised fine-tuning (SFT) or reinforcement learning (RL).
SFT-based methods train agents on human demonstrations or trajectories synthesized by stronger LLMs, providing dense action-level supervision~\cite{deng2023mind2web,zeng2023agenttuning,chen2023fireact,xi2024agentgym}.
However, collecting high-quality demonstrations can be expensive, and synthetic trajectories remain constrained by the coverage, quality, and biases of the teacher model or hand-designed workflow.
RL-based methods instead optimize agents using environment rewards or verifier feedback, allowing policies to improve through trial and error without step-by-step demonstrations~\cite{jin2025search,sun2025zerosearch,wang2026cooperative,li2026march,wu2026train,zhangprm2025}.
A central challenge in agentic RL is sparse and delayed rewards. Recent methods address this through improved credit assignment or auxiliary supervision. GIGPO, for example, estimates relative advantages at both the episode and step levels~\cite{feng2025group}.

\paragraph{World modeling and language-agent learning.}
World models have a long history in reinforcement learning, where they are trained on observed transitions to predict future states and rewards~\cite{ha2018worldmodels,hafner2020dreamer,hafner2021mastering}.
By learning environment dynamics, world models can support planning, improve sample efficiency, and provide richer learning signals.
Recent work extends world modeling to language agents, where states and observations are represented as text and transitions are induced by language actions.
RAP frames reasoning with language models as planning with an internal world model~\cite{hao2023rap}.
Dyna-style language-agent methods train action-conditioned transition models or align simulated next states with realized next states~\cite{lin2023dynalang,guo2025worldmodelling,yu2026rwml}.
Other work uses language-based simulators to support web-agent planning, verification, or data generation~\cite{gu2025webdreamer,chae2025webagentsworldmodels}.
Imagine-then-Plan uses a learned world model to generate imagined trajectories and adaptively choose the planning horizon~\cite{liu2026imaginethenplan}.
These studies demonstrate that predicting future observations can improve agent reasoning and decision making.

%% file: appendix/others.tex
\section{Licenses and Data Privacy}

\begin{table}[h]
\centering
\small
\begin{tabular}{p{0.20\linewidth}p{0.38\linewidth}p{0.22\linewidth}}
\toprule
\textbf{Artifact} & \textbf{Usage} & \textbf{License} \\
\midrule
verl-agent & RL training framework. & Apache License 2.0 \\
ALFWorld & Embodied agent environment. & MIT License \\
WebShop & Web-shopping agent environment. & MIT License \\
Search-R1 & Search/tool-use benchmark and data processing. & Apache License 2.0 \\
\bottomrule
\end{tabular}
\caption{Licenses of external artifacts used in this work.}
\label{tab:artifact-licenses}
\vspace{0.3em}

\end{table}

All models, datasets, and code artifacts used in this work are used in compliance with their respective licenses, and we retain the original attribution and license notices.
We verified that the dataset used in this work does not contain personally identifying information or offensive content. The data are fake environments, with no names, contact information, addresses, user identifiers, or other unique personal identifiers. Therefore, the issue of anonymizing personal information is not applicable to this study.

\section{Use of LLMs}

We used LLMs exclusively as writing assistants to refine language. In particular, their use was restricted to grammar correction, style improvement, and phrasing adjustments for clarity and conciseness.